\documentclass[10pt,twocolumn,letterpaper]{article}

 \usepackage{cvpr}              

\usepackage[linesnumbered,ruled,lined,commentsnumbered]{algorithm2e}
\usepackage{times}
\usepackage{helvet}
\usepackage{courier}
\usepackage[hyphens]{url}
\usepackage{graphicx} 
\usepackage{amsmath, xparse}
\usepackage{mathtools}
\usepackage[table,xcdraw]{xcolor}
\usepackage{color}
\usepackage{booktabs}
\usepackage{multirow}
\usepackage{adjustbox}
\usepackage{animate}
\usepackage{float}

\SetAlgoSkip{SkipBeforeAndAfter}

\newcommand\blfootnote[1]{%
  \begingroup
  \renewcommand\thefootnote{}\footnote{#1}%
  \addtocounter{footnote}{-1}%
  \endgroup
}

\definecolor{cvprblue}{rgb}{0.21,0.49,0.74}
\usepackage[pagebackref,breaklinks,colorlinks,citecolor=cvprblue]{hyperref}

\def\paperID{4322} 
\def\confName{CVPR}
\def\confYear{2024}

\title{Harnessing Meta-Learning \\for Improving Full-Frame Video Stabilization}

\author{
Muhammad Kashif Ali$^1$
~~~~~~~~
Eun Woo Im$^2$
~~~~~~~~
Dongjin Kim$^1$
~~~~~~~~
Tae Hyun Kim$^{1\dag}$\\
{\tt \{kashifali, iameuandyou, dongjinkim, taehyunkim\}@hanyang.ac.kr}\\
{$^1$Dept. of Computer Science, $^2$Dept. of Artificial Intelligence, Hanyang University}\\
}

\begin{document}
\maketitle
\begin{abstract}
Video stabilization is a longstanding computer vision problem, particularly pixel-level synthesis solutions for video stabilization which synthesize full frames add to the complexity of this task. These techniques aim to stabilize videos by synthesizing full frames while enhancing the stability of the considered video. 
This intensifies the complexity of the task due to the distinct mix of unique motion profiles and visual content present in each video sequence, making robust generalization with fixed parameters difficult. 
In our study, we introduce a novel approach to enhance the performance of pixel-level synthesis solutions for video stabilization by adapting these models to individual input video sequences. 
The proposed adaptation exploits low-level visual cues accessible during test-time to improve both the stability and quality of resulting videos. 
We highlight the efficacy of our methodology of ``test-time adaptation" through simple fine-tuning of one of these models, followed by significant stability gain via the integration of meta-learning techniques. 
Notably, significant improvement is achieved with only a single adaptation step. 
The versatility of the proposed algorithm is demonstrated by consistently improving the performance of various pixel-level synthesis models for video stabilization in real-world scenarios. \blfootnote{\hspace{-2em}$^\dag$: Corresponding author.}
\end{abstract}

\section{Introduction}
Today, the act of capturing and sharing visual content is deeply ingrained in our daily lives.
Millions of users rely on social networking platforms like YouTube and Facebook to document and share their favorite experiences with others.
However, the lack of specialized stabilization equipment, such as gimbals, often results in noticeably shaky and unstable videos.
This jitter affects the overall user experience and hinders effective visual communication. 
Consequently, the field of video stabilization has attracted considerable attention from both videographers and researchers alike, offering the potential to enhance the visual experience and support various downstream vision tasks.

Traditionally, video stabilization methods have followed a straightforward pipeline of motion estimation, smoothing, and compensation techniques involving spatial transformations.
Despite significant efforts to improve these transformation methods, the restoration process often comes at the expense of losing valuable visual content due to pixel projection, leading to irregular boundaries near the edges of stabilized videos.
To mitigate this issue, cropping is commonly employed, resulting in loss of visual resolution. 
However, recent advances in deep learning methodologies have brought new possibilities for content preservation on the cropped region.
Approaches such as inpainting the missing regions~\cite{OVS, choi2021self} or defining an end-to-end pipeline that simultaneously stabilizes and synthesizes missing regions~\cite{bmvc_ver, DIF, liu2021hybrid} offer promising solutions.
However, achieving end-to-end feed-forward pixel-level stabilization remains challenging due to the inherent difficulty of this task and the diverse scenarios in real-world video.

Notably, the pioneering works of Choi~\etal~\cite{DIF} and Ali~\etal~\cite{bmvc_ver} have initiated the exploration of end-to-end full-frame video stabilization methods.
Choi~\etal~\cite{DIF} introduced an optical flow-based frame interpolation method (termed DIFRINT) that stabilizes videos through multiple temporal interpolations.
On the other hand, Ali~\etal~\cite{bmvc_ver} proposed Deep Motion-Blind Video Stabilization (DMBVS), a feed-forward method, which is trained on a dataset that consists of stable and unstable videos with similar perspectives.
Despite their contributions, both methods face certain limitation,
for instance, DIFRINT encounters challenges in preserving perceptual quality over multiple interpolation iterations and is prone to temporal artifacts near the motion boundaries where occlusion and dis-occlusion occur.
Conversely, DMBVS generates visually appealing frames but lacks a mechanism to control the level of stability in the resulting videos.

To overcome these limitations, one potential approach is to make these models adaptive and leverage the spatiotemporal cues present in specific scenes, similar to the strategies employed by classical approaches based on spatial transformations.
However, a shortcoming of test-time adaptation in neural approaches is the considerable time and resources required to adapt to new data. 
This can be alleviated by employing techniques investigated in meta-learning literature, as similar techniques have been proven effective in various computer vision tasks such as video super-resolution~\cite{lee2021dynavsr}, frame interpolation~\cite{choi2021test}, and visual tracking methods~\cite{metatrack}. 
We hypothesize that these techniques can also improve video stabilization approaches by quickly adapting to the input data at test time without using the ground truth stable data.
Using these techniques, we can combine the strengths of deep learning methods, which provide superior quality, with classical methods that provide better stability, along with the added benefit of giving users more control over the stability and quality of the resulting videos.

In this work, we propose a scene-adaptive video stabilization method that can quickly adapt to unseen videos at test time. 
At test time, we improve both the picture quality and stability of full-frame video stabilization models. 
To the best of our knowledge, this is the first integration of meta-learning in the field of video stabilization.
The proposed fast adaptation algorithm can be seamlessly integrated with any off-the-shelf end-to-end pixel synthesis stabilization models.
Additionally, it allows the adapted models to achieve an $\sim$ 8\% absolute gain in stability and provides state-of-the-art results for pixel synthesis methods for video stabilization. 

\noindent We summarize our contributions as follows:\\
$\bullet$ We integrate the meta-learning algorithm, which improves the performance of full-frame video stabilization models by adapting model parameters to various scenes with distinct motion profiles and content.\\
$\bullet$ Our method equips these fixed-performance models with a moderate control mechanism for various aspects of video stabilization and consistently improves the performance in these aspects by increasing the number of adaptation steps.\\
$\bullet$ We achieve SOTA video stabilization results on the evaluation datasets and our method outperforms the long-standing SOTA methods for this task.

\section{Related  works}
This section summarizes the related literature on video stabilization and meta-learning for computer vision tasks.
\subsection{Video stabilization}
Conventionally, video stabilization approaches can be classified into three distinct categories, 3D, 2.5D and 2D approaches.
The 3D approaches for video stabilization model the camera trajectories in the 3D space. Various techniques such as depth information~\cite{liu2012video}, gyroscopic data~\cite{karpenko2011digital} structure from motion~\cite{liu2009content}, light fields~\cite{smith2009light}, and 3D plane constraints~\cite{zhou2013plane} have been used to stabilize videos in 3D space.
Despite their ingenious formulations, these approaches face difficulties in handling dynamic scenes containing multiple moving objects; therefore, 2D approaches which limit their scope to spatial transformations like homography and affine transformations became the tool of choice for researchers.
Generally, these approaches track and stabilize the trajectories of prominent features. Doing so introduces loss of visual content near the frame boundaries which is often concealed by cropping and up-scaling the resultant video.

For 2D stabilization, Buehler~\etal~\cite{buehler2001non} estimated camera poses in shaky videos and used non-metric image-based rendering to stabilize videos.
Matsushita~\etal~\cite{matsushita2006full} estimated simplistic 2D global transformations to warp the unstable frames to produce stable video, and Liu~\etal~\cite{liu2013bundled} extended this phenomenon to grid-based warping for smoothing feature trajectories.
Grundmann~\etal~\cite{grundmann2011auto} presented an $L_1$-based objective function for estimating stable camera trajectories, whereas Liu~\etal~\cite{liu2011subspace} utilized eigen-trajectory smoothing for this task.
Goldstein~\etal~\cite{goldstein2012video}, Lee~\etal~\cite{lee2009video}, and Wang~\etal~\cite{wang2013spatially} employed epipolar geometry-based optimization models for stabilizing videos.

Inspired by these approaches and looking at their shortcomings in handling the independent motion of multiple objects, Liu~\etal~\cite{liu2013bundled} highlighted the importance of ``\emph{relatively}" denser inter-frame motion through optical flow for video stabilization.
Their findings inspired most of the modern video stabilization methodologies that are currently being used professionally to this day in apps like Blink, Adobe Premiere Pro, and Deshaker. 
Many recent works~\cite{OVS, DUT, DIF, yu2019robust, yu2020learning, liu2021hybrid} rely on optical flow as an irreplaceable backbone for the definition of their approaches. Geo et al.\cite{geo2023globalflownet} further improved on these methods and finetuned a conventional flow estimation network to estimate only the camera motion component of optical flow (termed global optical flow) and used it to define warping fields for video stabilization.
Please note that, unlike the conventional deep stabilization methods, Ali~\etal~\cite{bmvc_ver} highlighted the importance of perspective in training data and the power of traditional deep convolutional neural networks (CNNs) by learning to synthesize stable frames entirely through learned implicit motion compensation from neighboring frames, and Choi~\etal~\cite{DIF} proposed an iterative interpolation strategy for stabilizing videos.
Please note that these two methods are the only proposed methods for pixel synthesis end-to-end full-frame video stabilization.

\subsection{Meta learning and test-time optimization}
For deep video stabilization methods, some literature has been investigated on test-time adaptation inspired by the conventional optimization approaches.
Yu~\etal~\cite{yu2019robust} proposed to stabilize videos by optimizing the motion vector warp field in CNN weight-space. Liu~\etal~\cite{liu2021hybrid} propose to learn radiance fields for distinct scenes, and Xu~\etal~\cite{DUT} defined a pipeline inspired by~\cite{grundmann2011auto, liu2013bundled} with the help of a modular pipeline catering to estimating and iteratively smoothing the motion trajectories and reprojecting the unstable frames to follow a smooth global motion profile. Despite the ingenuity of these approaches, these methods significantly hamper the time required for stabilizing videos. 

Contrary to the conventional optimization-based video stabilization approaches, we aim to investigate faster test-time adaptability for full-frame video stabilization approaches inspired by its
recent success in various computer vision tasks such as video super-resolution~\cite{gupta2021ada, lee2021dynavsr}, visual tracking~\cite{metatrack}, video segmentation~\cite{behl2020meta}, object detection~\cite{deng2021minet}, human pose estimation~\cite{cho2021camera}, image enhancement~\cite{ma2023bilevel}, and video frame interpolation~\cite{metavfi}.
Typically, meta-learning algorithms can be categorized into three main groups: metric-based, network-based, and optimization (or gradient)-based algorithms.
From the optimization-based category of meta-learning, model agnostic meta-learning (MAML)~\cite{maml} has become the tool of choice for researchers investigating computer vision tasks~\cite{cheng2021meta, fu2019embodied, hosseinzadeh2023few, lee2019metapix, lin2021self, lu2020deep, ren2020video, wang2021meta, wang2020tracking, yang2023toward, zhao2021one, zou20202} due to its effectiveness, generalizability, and simplicity.

In light of recent literature, and its success in low-level computer vision tasks, we investigate the applicability of this technique for pixel-level synthesis solutions for video stabilization and propose a new algorithm that combines the strengths of conventional spatial transformation-guided video stabilization approaches and regressive properties of pixel-level synthesis video stabilization approaches.
The proposed algorithm allows the parameters of the feed-forward video stabilization models to be updated quickly with respect to the unique motion profiles and diverse image content present in each scene and allows the adapted model to stabilize extremely shaky videos while preserving visual quality and resolution.
The proposed model also provides the user with the ability to control the level of stability and quality preservation (up to a certain degree); which is unattainable with currently available regressive solutions for this task.

\section{Proposed method}
This section begins by presenting the problem setup of pixel-level regressive video stabilization. Next, we discuss the proposed algorithm, outline the meta-training objective functions, and discuss the inference strategy.

\begin{figure}[!t]
\vspace{-10px}
\begin{center}
\begin{frame}{}
\animategraphics[autoplay,loop,width=0.90\linewidth]{1}{figures/dif_ali/dif_figure_ver5_}{1}{3}
\\[-1.5em]
\end{frame}
\end{center}
\caption{\textbf{Recurrence related artifacts.} Wobble artifacts observed in the frame recurrent settings for full-frame video stabilization models. Please note that this figure includes animated content and is best viewed on a computer with Adobe PDF Reader.}
\vspace{-1.5em}
\label{temporal_artifacts}
\end{figure} 

\subsection{Problem set-up}
Consider an unstable video containing $n$ frames as $V$ = \{$I_{0}$, $I_{1}$, ..., $I_{n}$\}. 
The goal of the video stabilization methods is to predict a stable video $\hat{V}$ = \{$\hat{I_{0}}$, $\hat{I_{1}}$, ..., $\hat{I_{n}}$\} using a stabilization network $f_{\theta}$ given the unstable input video $V$,
and the predicted video $\hat{V}$ contains similar content to $V$ with a stabilized camera trajectory.
Conventionally, stabilization methods based on pixel synthesis~\cite{bmvc_ver, DIF} employ a sliding window strategy that considers a local temporal window containing $2k + 1$ frames ($\{I_{t - k}, ..., I_{t}, ..., I_{t + k}\}$) and produce a stabilized frame $\hat{I_{t}}$ as:
\begin{equation}
    \hat{I_{t}} = f_{\theta}({S}_t),
\label{eq:frame_ip_op_non_rec}
\end{equation}
where $S_t$ denotes the local temporal window of $2k+1$ consecutive frames. 
This temporal window strategy allows the model to regress missing information in synthesized stable frames. 
For instance, temporal window of $5$ consecutive unstable frames (\ie $S_t=\{I_{t - 2},I_{t - 1}, I_{t}, I_{t + 1},I_{t + 2}\}$) is used in DMBVS, 
and a temporal window of $3$ consecutive frames with frame recurrence (\ie $S_t = \{\hat{I}_{t - 1}, I_{t}, I_{t + 1}\}$) is utilized in DIFRINT.
Note that, the initial $k$ and last $k$ frames cannot be stabilized with window-based approaches, but we use $0 \le t \le T$ for notational simplicity throughout this paper.

These pixel-synthesis methods are straightforward and allow for end-to-end learning and inference. 
However, one of the main drawbacks of these works is the limited performance in terms of stability. 
While the frame recurrence schemes can improve the stability of these methods by propagating synthesized content to regress future frames and can be used with any window-based approach, 
these approaches can also compromise the quality and introduce wobble (jitter) artifacts, as shown in Fig.~\ref{temporal_artifacts}.
Despite the limited performance in stabilization, pixel-level synthesis solutions are still promising, 
because they can easily produce full-frame videos after stabilization. Therefore, we formulate our fast adaptation method based on these pixel-level synthesis approaches to improve both stability and image quality.

\begin{figure}[t]
    \centering
    \includegraphics[width=0.9\linewidth]{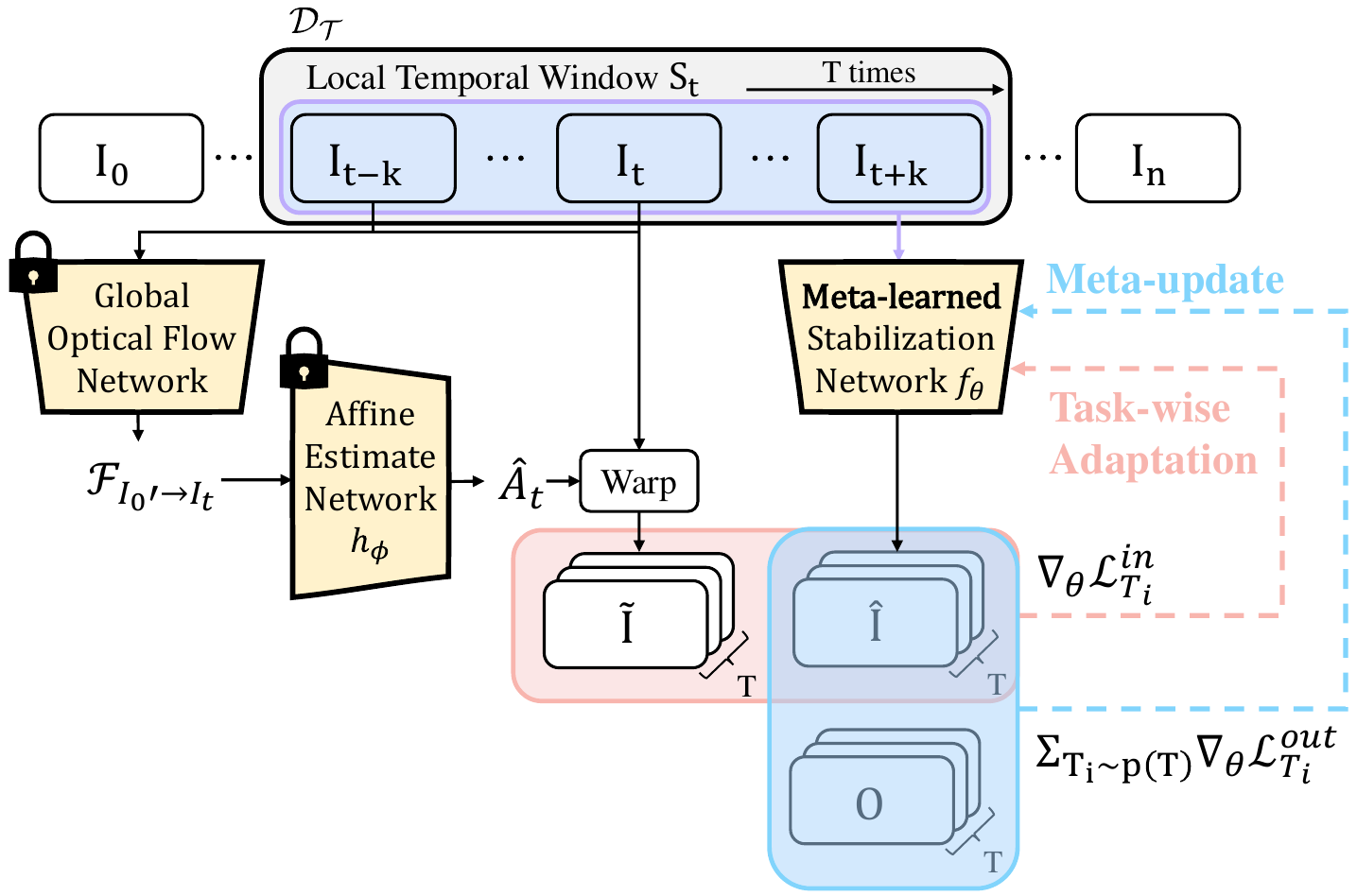}
    \caption{\textbf{Overview of the proposed meta-training process.} This figure illustrates the overall pipeline of the training process. The model in the inner loop gets a sequence of local temporal windows ($S_t \in \mathcal{D_{T}}$) and synthesizes stable frames. The synthesized frames are penalized according to the aligned frames in the inner loop. For the outer loop, the deviation of synthesized frames is measured with the corresponding DeepStab~\cite{wang2018deep} stable frames. At inference time, only the inner loop optimization is needed.}
    \label{overview_fig}
    \vspace{-20pt}
\end{figure}

\subsection{Meta-learning for video stabilization} \label{main_method}
Our key observation highlights the challenge that pixel-level synthesis stabilization models face when dealing with motion in specific scenarios. 
This challenge arises from biases in conventional training data and the complexities associated with using motion cues from raw pixel values.
Therefore, we hypothesize that in real-world videos, the motion profiles can vary significantly even within the same video content, for which models with fixed parameters might be ineffective; thus, to make these models more effective, we propose a fast test-time adaptation strategy that allows these models to explicitly look for and utilize visual cues for specific unique scenarios for better compensation of camera shakes. 
Specifically, to aid the adaptation process, we use MAML~\cite{maml}, which is known for its ability to effectively adapt to new tasks. 
The MAML algorithm consists of two components: an inner loop and an outer loop.
Within the inner loop, the parameters of the models are adapted through a small number of adaptation steps for each specified task. 
Following this adaptation, in the outer loop, test sets for the task in the inner loop are sampled to evaluate the generalization of the adapted model.
In this work, to define a scene-adaptive video stabilization approach, we consider a short sequence of frames as a ``\emph{task}''; which is then used for fast adaptation to unseen videos through the proposed algorithm.
We employ a feed-forward video stabilization network $f_{\theta}$, which takes a set of $\emph{2k + 1}$ neighboring frames as in Eq.~\ref{eq:frame_ip_op_non_rec} to synthesize its stable counterpart $\hat{I}_{t}$,
and we use the DMBVS and DIFRINT as our baselines.
The task in our formulation is defined as the minimization of both of the aforementioned objectives in the MAML framework on $\emph{T}$ consecutive input frame sequences from unstable videos. 
The overall process of our proposed meta-training process is illustrated in Fig.~\ref{overview_fig}.

During the training-phase, each task $\mathcal{T}_i$ is sampled from the DeepStab dataset~\cite{wang2018deep} ($\mathcal{D}_{\mathcal{T}_i}$).
The inner loop update is governed with the help of an inner loop loss function $\mathcal{L}^{\text{in}}_{\mathcal{T}_i}$ which does not require the ground truth counterpart (as shown in the Fig.~\ref{overview_fig}), whereas, the parameter update at the meta-stage (outer loop) is governed by $\mathcal{L}^{\text{out}}_{\mathcal{T}_i}$ for which we utilize the stable videos from the same dataset. 
In our formulation, the inner loop loss is focused on input-specific information available at test time which can be used to improve both stability and perceptual quality, whereas, the outer loop loss focuses more on visual quality to instill a sense of mitigating jerk-related degradations such as blur and distortions, hence it requires the stable counterparts of the DeepStab~\cite{wang2018deep} videos; thus, meta-learning is employed to take into consideration both the input specific cues at test time while making the models under consideration stronger in each of the concerned aspect of video stabilization. It is worth noting that despite focusing more on one aspect, both the discussed losses contain parts that penalize deviation from other aspects as well. 
\vspace{-10pt}
\subsubsection{Objective functions}
Ali~\etal~\cite{bmvc_ver} showed that various motion-related objectives can be abstracted in pixel space, therefore, we implicitly define our motion penalties in both pixel space with the help of a rigid transform estimation module and optical flow space, as there is no ground-truth available for video stabilization, and the videos in the DeepStab dataset~\cite{wang2018deep} contain perspective mismatch~\cite{bmvc_ver}. We intentionally opt for rigid transforms in our formulation, as these transforms do not consider scale and shear change, which often causes visual distortions in the transformed images. 
These unique properties of rigid transforms not only govern the stabilization process but also limit the deviation of visual content from that of actual content as the transformed images are wobble-free.
We will now elaborate on the details of our rigid transform regression module and then define the formulation of the proposed losses $\mathcal{L}^{\text{in}}_{\mathcal{T}_i}$ and $\mathcal{L}^{\text{out}}_{\mathcal{T}_i}$, and the proposed algorithm.

First, for rigid transform estimation, we separately trained and froze our affine motion estimation network $h_{\phi}$. 
This network $h_{\phi}$ is pre-trained with the 
global optical flow $\mathcal{F}_{I \rightarrow I'}$ (as presented in~\cite{geo2023globalflownet}) estimated between randomly transformed images $I$ and $I'$ with rigid transforms to regress rotation and translation parameters of the rigid affine transform.
We use the global optical flow instead of a conventional optical flow as the input of our $h_{\phi}$ network since it masks the flow of dynamic objects from the evaluated flow and is also robust against crops in the input images, which aids the proposed rigid transform estimation network to focus on removing camera shake in a video rather than local motion. 
To be specific, the proposed network regresses the rigid affine transform parameters as follows:
\begin{equation}
\hat{\mathcal{A}}_{I'} =  h_{\phi} (\mathcal{F}_{I \rightarrow I'}),
\end{equation}
where $\hat{\mathcal{A}}_{I'}$ denotes the estimated rigid transform, and $h_{\phi}$ is the proposed affine estimation network which renders rotational and translational parameters of the rigid transformation $\hat{\mathcal{A}_{I'}}$ from the global optical flow ($\mathcal{F}_{I \rightarrow I'}$) between the frames $I$ and $I'$. Then, our $h_{\phi}$ network can be used to align short sequences of input frames by estimating transformation parameters w.r.t. the first input frame as follows:
\begin{figure}[t]
    \centering
    \includegraphics[width=0.98\linewidth]{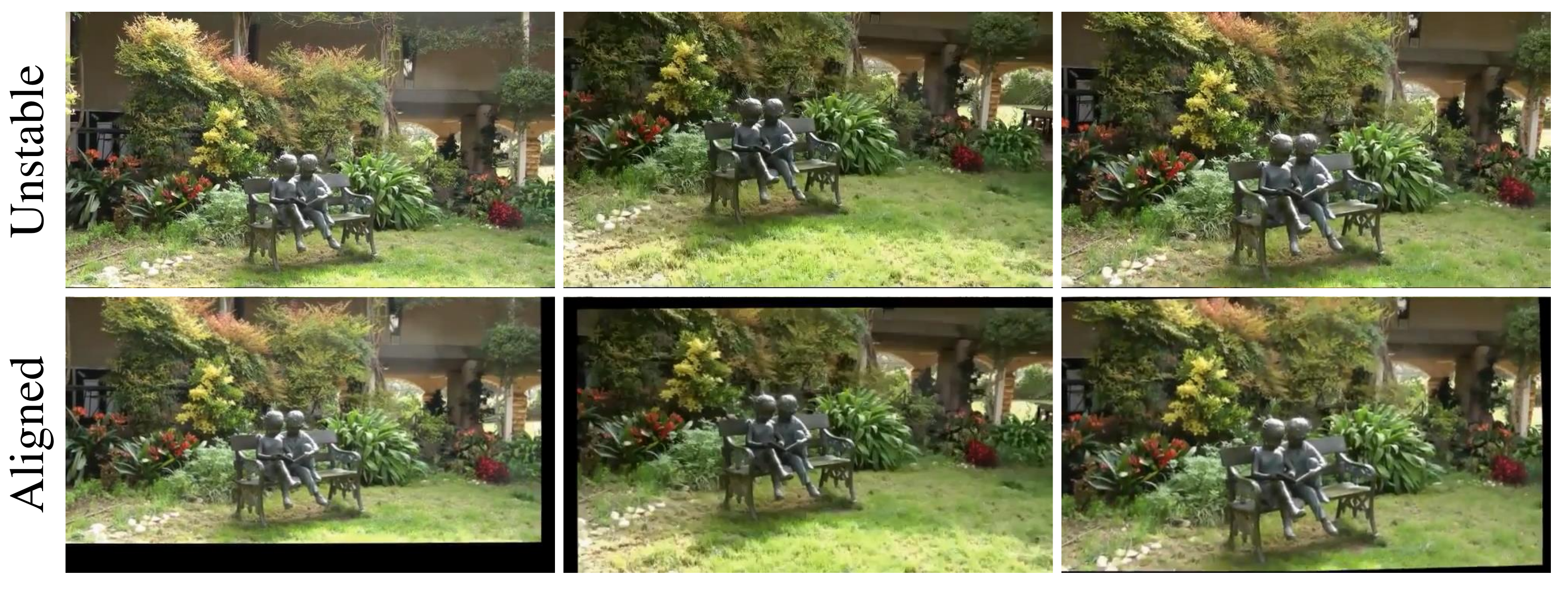}
    \caption{\textbf{Affine alignment.} This affine alignment strategy is analogous to the classical stabilization strategies which estimate and smooth transforms to stabilize videos. Please note that these frames are not neighboring frames and were selected to highlight the crops near the image boundaries in aligned frames $\tilde{V}$.}
    \label{coarse_stable_frame}
    \vspace{-15pt}
\end{figure}
\begin{equation}
\begin{split}    
    &\mathcal{\hat{A}}_{t} = h_{\phi} (\mathcal{F}_{I_{0'} \rightarrow I_{t}}),~t \in \{1, ..., T\},\\
    &\tilde{I}_{t} = \mathcal{W}(I_{t}, \mathcal{\hat{A}}_{t})~, ~\tilde{V} = \{\tilde{I}_{0'}, \tilde{I}_{1}, ..., \tilde{I}_{T}\}.
\end{split}
\label{eq:8}
\end{equation}
Here, $\mathcal{\hat{A}}_{t}$ denotes the estimated rigid transform that aligns frame $I_t$ to the first frame ($I_{0'}$) of the sequence, $T$ denotes the number of consecutive frames, $\mathcal{W}$ represents the spatial warp operator, and $\tilde{I}_t$ refers to the warped frame. Please note that ${I_{0'}}$ denotes the first frame of the sampled short sequence instead of the actual first frame of the video.
The set ($\tilde{V}$) indicates the aligned frames. 
Note that($I_{0'}$) is used as the reference frame, so alignment is not required, but $\tilde{I}_{0'}$ is used to keep the notation consistent. 

These aligned frames can be used as a stabilization guide for the proposed algorithm, but these frames include significant cropped regions near the image boundaries as shown in Fig.~\ref{coarse_stable_frame}; thus, these frames cannot be used directly as ground-truth stable frames like the ones used in DMBVS. 
Therefore, we define our inner loop loss for meta-learning as the sum of global camera motion and perceptual distance between these aligned frames and the regressed frames from the feed-forward stabilization networks $f_{\theta}$ as follows:
\vspace{-5pt}
\begin{equation}
    \mathcal{L}^{\text{in}}_{\mathcal{T}_i} = \lambda_s \cdot {\mathcal{L}^{\text{in}}}_{\text{stability}} + \lambda_p \cdot {\mathcal{L}^{\text{in}}}_{\text{quality}},
\label{eq:9}
\end{equation}
where $\lambda_s$ and $\lambda_p$ are associated weights for stability and quality loss, respectively.
The inner loop stability loss (${\mathcal{L}^{\text{in}}}_{\text{stability}}$) is defined as the absolute mean of global optical flow between the regressed frame $\hat{I}_t$ and the rigid-affine aligned frame $\tilde{I}_t$ as:
\vspace{-13pt}
\begin{equation}
    {\mathcal{L}^{\text{in}}}_{\text{stability}} = \sum_{t = 1}^T \frac{1}{N} \sum_{N} | \mathcal{F}_{\hat{I}_t \rightarrow \tilde{I}_t} |.
\end{equation}
Here, $N$ represents the total number of pixels in the regressed frame. Please note that the employed global optical flow estimation network is quite robust against augments that resemble the cropped regions in the warped frames $\tilde{I}_t$ and fills these holes by utilizing the visual context from the input images\footnote{Please refer to the supplementary material for robustness comparison of the employed and a conventional optical flow estimation network.}.
The intuition behind this loss formulation is to enforce dense alignment between the regressed and aligned sequences, as, ideally, the regressed frames and the aligned frames should align perfectly.
However, this loss by itself cannot justify the synthesis of legible content, as there can exist multiple solutions to the optical flow equation~\cite{btcc_work}; therefore, strong visual penalties should be introduced to ensure content preservation.
We introduce these penalties in the form of perceptual loss~\cite{johnson2016perceptual}, a contextual loss, and a feature-based gram matrix loss to preserve the visual content and style of the input videos. Please note that throughout our experiments, we fix $T = 5$ due to resource limitations.
The proposed loss to secure video quality is defined as:
\vspace{-10px}
\begin{equation}
\begin{split}
    {\mathcal{L}^{\text{in}}}_{\text{quality}} = &\sum_{t=0}^{T}\sum_{l}\left\|\phi_{l}\left(\hat{I}_{t}\right)-\phi_{l}\left(\tilde{I}_{t}\right)\right\|^2_{2} \\
    & + \sum_{t=0}^{T}\sum_{l}\left\|G(\phi_{l}\left(\hat{I}_{t}\right))-G(\phi_{l}\left(\tilde{I}_{t}\right))\right\|^2_{2}\\
    & - \log(CX( \phi _ l (\hat{I}_{t})  , \phi _ l (\tilde{I}_{t}))).
\end{split}
\label{eq:in_qual_loss}
\end{equation}
Here $\phi_{l}(\cdot)$ represents layers of a VGG-16 network till the layer $relu\_4\_3$ (trained on the ImageNet dataset~\cite{deng2009imagenet}). $G$ represents the gram matrix of features extracted from the corresponding layer $l$ and $CX(\cdot)$ represents contextual loss.
We employ the contextual and perceptual losses in our formulation in line with the previous literature~\cite{bmvc_ver}, which has shown the effectiveness of these losses for video stabilization.
In particular, the addition of gram matrix loss further encourages the models to synthesize realistic frames.

The combination of both 
of these losses is used to carry out the inner loop update of the proposed algorithm to obtain the adapted network parameter $\theta'_i$.
Please note that this inner loop update step can be repeated $M$ times. 

Next, within the outer loop, our network parameters are updated to minimize the different stability and quality penalties for $f_{\theta'_{i}}$ w.r.t. $\theta$ on different sampled frame sequences along with their stable counterparts from the DeepStab dataset~\cite{wang2018deep}. In the outer loop update, we focus more on the qualitative objectives due to the availability of stable videos which contain roughly the same content with better quality as compared to the unstable videos.

The motion loss for the outer loop update is defined as the deviation between the global camera motion of synthesized frames and their stable counterparts as: 
\begin{equation}
    {\mathcal{L}^{\text{out}}}_{\text{stability}} = \sum_{t = 0}^{T-1} \frac{1}{N} \sum_{N} \left\| \mathcal{F}_{\hat{I}_t \rightarrow \hat{I}_{t + 1}} - \mathcal{F}_{O_t \rightarrow {O}_{t + 1}} \right\|^2_{2},
\end{equation}
where $O_t$ represents the target stable frame in the DeepStab dataset corresponding to the predicted stable frame $\hat{I}_t$.
This loss further enforces the learned stability of the model under consideration with smooth real-world trajectories.
Similar to the stability loss in the inner loop, this loss alone cannot justify the preservation of legible content; therefore, a qualitative penalty is also added in the outer loop update.

Since both the stable and unstable videos in the DeepStab dataset contain large disjoint perspectives~\cite{bmvc_ver}, a non-local criterion is needed for a quality guidance. 
We take inspiration from Ali~\etal~\cite{bmvc_ver} to define our non-local quality penalty using contextual loss~\cite{mechrez2018contextual}, which compares unaligned image regions with similar semantics and has been shown to be useful in improving the quality of synthesized stable frames~\cite{bmvc_ver}.
The outer loop quality loss with the ground-truth target $O_t$ is defined as:
\begin{equation}
    {\mathcal{L}^{\text{out}}}_{\text{quality}} = -\log(CX( \phi ^ l ({\hat{I}}_{t})  , \phi ^ l ({O}_{t}))),
\end{equation}
and the final loss for the outer update is defined as:
\begin{equation}
   \mathcal{L}^{\text{out}}_{\mathcal{T}_i} = {\mathcal{L}^{\text{out}}}_{\text{stability}} 
 + {\mathcal{L}^{\text{out}}}_{\text{quality}}.
 \label{eq:14}
\end{equation}
\vspace{-20pt}
\subsubsection{Meta-training and inference}
\begin{algorithm}[!t]
    \small
    \LinesNumbered
    \SetKwInOut{Require}{Require}
    
    \Require{uniform distribution over sequences $p(\mathcal{T})$, adaptation number $M$, learning rate $\alpha$, $\beta$}
    \BlankLine

    \While{not converged}{
        Initialize parameters $\theta_i \leftarrow \theta$\;
        Sample batch of sequences $\mathcal{T}_i \sim p(\mathcal{T})$\;
        \ForEach{i}{
            Sample local temporal windows $\mathcal{D}_{\mathcal{T}_i} = \{ S_0, S_1, ..., S_t\}$ from $\mathcal{T}_i$\;
            \For{$m \gets 1$ to $M$} {
                Compute $\hat{\mathbf{V}}$, $\tilde{\mathbf{V}}$ in Eq.~\eqref{eq:frame_ip_op_non_rec},~\eqref{eq:8}\;
                Evaluate $\nabla_{\theta_i} \mathcal{L}^{\text{in}}_{\mathcal{T}_i}(f_{\theta_i})$ using $\mathcal{L}_{\mathcal{T}_i}$ in Eq.~\eqref{eq:9}\;
                ${\theta_i}' = \theta_i - \alpha\nabla_{\theta_i}\mathcal{L}^{\text{in}}_{\mathcal{T}_i}(f_{\theta_i})$\;
            }
        }
        Sample $\mathcal{D}'_{\mathcal{T}_i} = \{ (S_0,O_0), (S_1,O_1), ..., (S_t,O_t)\}$ from $\mathcal{T}_i$ for meta-update\;
         $\theta \leftarrow \theta - \beta\nabla_{\theta}\sum_{\mathcal{T}_i \sim p(\mathcal{T})}\mathcal{L}^{\text{out}}_{\mathcal{T}_i}(f_{{\theta_i}'})$ using each $\mathcal{D}_{\mathcal{T}_i}'$\;
        
    }
    \caption{Meta-Training.} 
    \label{alg:meta_training}
\end{algorithm}
The overall training algorithm is presented in Alg.~\ref{alg:meta_training}.
Please note that at the test-time, only the inner loop loss is needed to update the meta-trained parameters and the updated parameters are used to synthesize the final stabilized results in a feed-forward manner.
It is worth mentioning that we experimented with a fixed number of adaptation iterations and a patch size of $320 \times 320$ during the inference time to further expedite the adaptation process and empirically found that even with as low as $100$ adaptation iterations on randomly sampled sequences from the test videos, the meta-trained models adapt quite well due to the similarity in motion profiles and the content of the videos.
This process significantly cuts down the adaptation time as most of the videos from the evaluation dataset~\cite{liu2013bundled} contain over $700$ frames.
Our fast adaptation algorithm is presented in Alg.~\ref{alg:meta_inference}.
Please refer to the accompanied supplemental for a detailed description of the implementation details and experiments.
\vspace{-10pt}
\begin{algorithm}[!t]
    \small
    \SetAlgoLined
    \SetKwInOut{Require}{Require}
    \Require{meta-trained model $f_{\theta}$, test sequence $\mathcal{T}$, adaptation number $M$, learning rate $\alpha$}
    \BlankLine
    
    Construct local temporal windows $\mathcal{D}_{\mathcal{T}} = \{ S_0, S_1, ..., S_t\}$ from $\mathcal{T}$\;
    \For{$m \gets 1$ to $M$} {
        Compute $\hat{\mathbf{V}}$, $\tilde{\mathbf{V}}$ in Eq.~\eqref{eq:frame_ip_op_non_rec},~\eqref{eq:8}\;
        Evaluate $\nabla_{\theta} \mathcal{L}^{\text{in}}_{\mathcal{T}}(f_\theta)$ using $\mathcal{L}_{\mathcal{T}}$ in Eq.~\eqref{eq:9}\;
        $\theta' = \theta - \alpha\nabla_{\theta}\mathcal{L}^{\text{in}}_{\mathcal{T}}(f_\theta)$\;
    }
    Stabilize video $\hat{\mathbf{V}} = f_{\theta'}(\mathbf{V})$ with sliding window strategy in Eq. (\ref{eq:frame_ip_op_non_rec})\;
    \Return stabilized video $\hat{\mathbf{V}}$
    \caption{Meta-Inference.}
    \label{alg:meta_inference}
\end{algorithm}
\vspace{10pt}
\section{Ablation study} \label{ablation_section}
To properly evaluate the efficacy of each of the modules and objective functions, we conducted thorough ablation studies and present our findings below.
We first present the contribution of each of the losses presented and then present the category-specific hyperparameters in this section.\\
\textbf{Objective function contribution.}
We explore the influence of each loss term presented in Eq.~\ref{eq:9} from the main paper (${\mathcal{L}^{\text{in}}}_{\text{quality}}$ and ${\mathcal{L}^{\text{in}}}_{\text{stability}}$) concerning different weights of each loss term in the adaptation process. 
\begin{figure}[t]
    \centering
    \begin{subfigure}[b]{1\linewidth}
        \centering
        \caption{Stability}
        \includegraphics[width=0.9\linewidth]{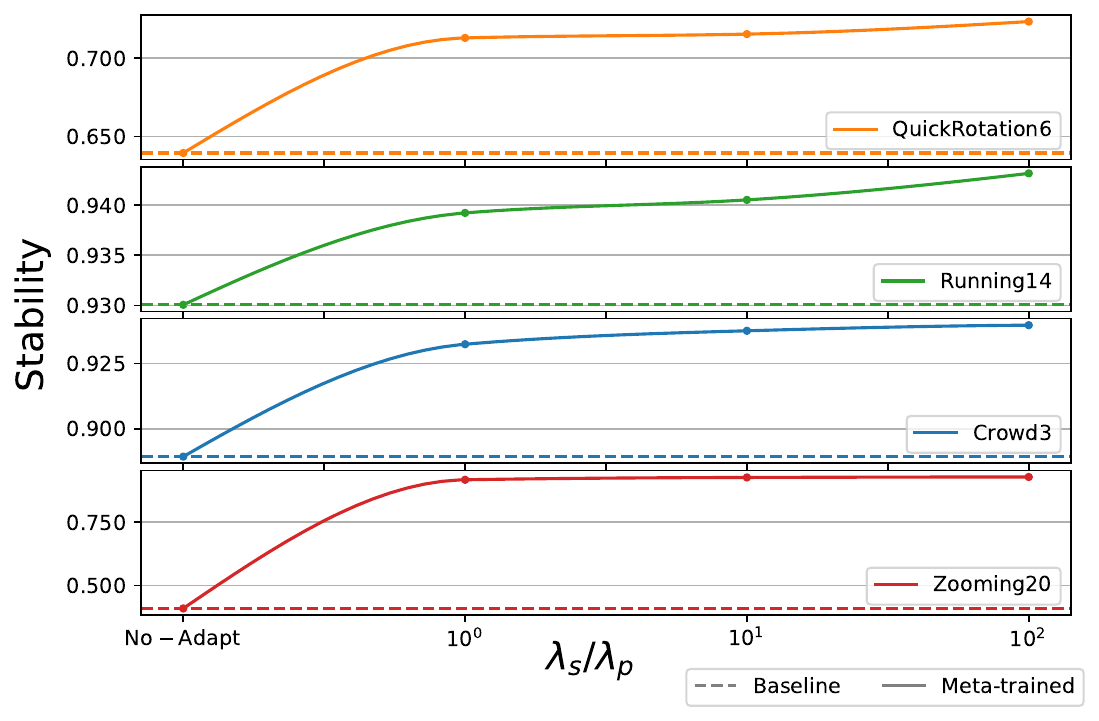}
        \label{fig:stability_ablation}
    \end{subfigure}
    \\[-0.6em]
    \begin{subfigure}[b]{1\linewidth}
        \centering
        \caption{Distortion}
        \includegraphics[width=0.9\linewidth]{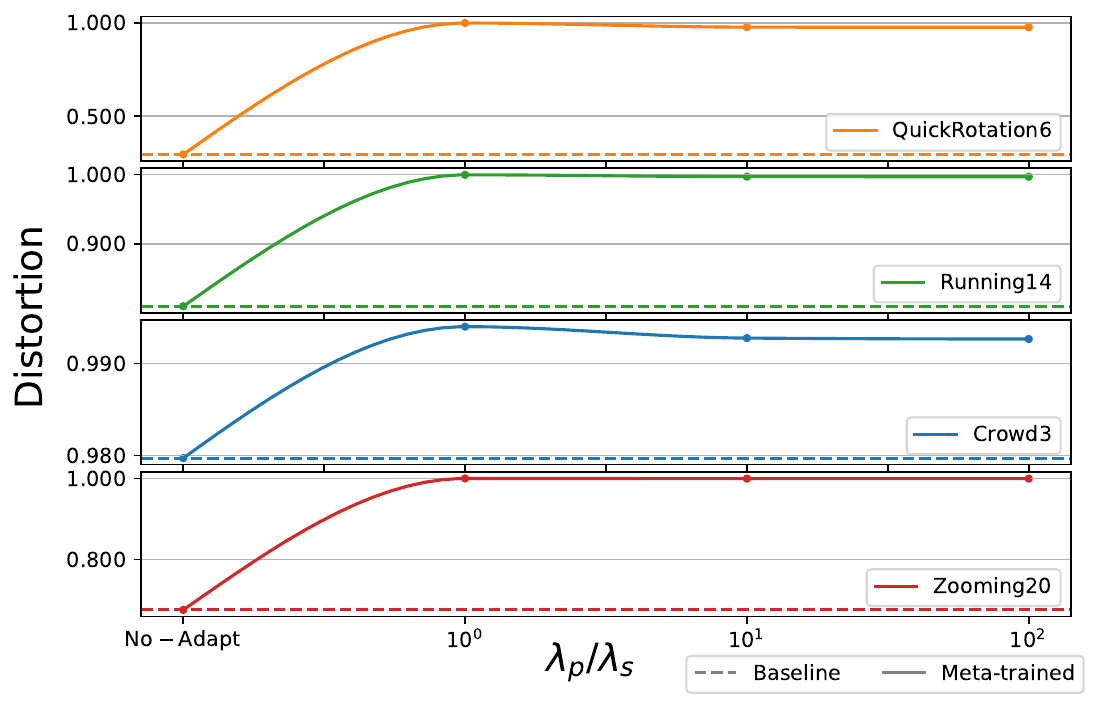}
        \label{fig:distortion_ablation}
    \end{subfigure}
    \vspace{-10pt}
    \caption{\textbf{Contribution of each objective function.}
    a) The effects of stability loss during the adaptation stage.
    A higher weight for the proposed stability loss positively affects the stability score.
    b) The effects of quality loss during the adaptation stage.
    A higher weight for quality loss positively affects the distortion score.}
    \label{fig:stab_and_dist_ablation}
    \vspace{-18pt}
\end{figure}

To properly ablate the contribution of each of the proposed losses, we randomly sample 4 videos from the NUS dataset~\cite{liu2013bundled} and repeat the adaptation process with various ratios of $\lambda_s$ and $\lambda_p$, and present our findings in Fig.~\ref{fig:stab_and_dist_ablation}.
For our ablation studies, we choose the meta-trained DMBVS~\cite{bmvc_ver}.
Note that similar phenomenons were observed with the meta-trained DIFRINT~\cite{DIF}, therefore, we only present the findings from one of the considered models in Fig.~\ref{fig:stab_and_dist_ablation}.
It is evident from Fig.~\ref{fig:stability_ablation}, that increasing the weights for the proposed stability loss positively affects the stability of the resultant videos and an increasing trend is observed in terms of stability metric results.
As for the quality loss, a similar increasing trend for distortion score is observed as evident from Fig.~\ref{fig:distortion_ablation}. Please note that the presented results in the main manuscript and this supplemental were generated with a \emph{10:1} ratio of $\lambda_s$ and $\lambda_p$.
\newline
\noindent
\textbf{Category-specific ratios.}
Each video category within the NUS dataset~\cite{liu2013bundled} exhibits distinct characteristics, necessitating tailored weighing configurations to achieve optimal results.
This subsection presents the findings of our study for the category-specific hyperparameters on individual video categories.
Please note that the presented results (in both the main paper and this supplemental) were evaluated on hyperparameters that demonstrated optimal performance across all the video categories.
However, we found the performance on distinct motion profiles can be further improved (by 1$\sim$2\%) by selecting specific weights for the stability and quality losses during the adaptation process. We present the category-specific weights in Tab.~\ref{tab:cat_spec_weights}. 

\begin{figure}[t]
    \centering
    \includegraphics[width=0.98\linewidth]{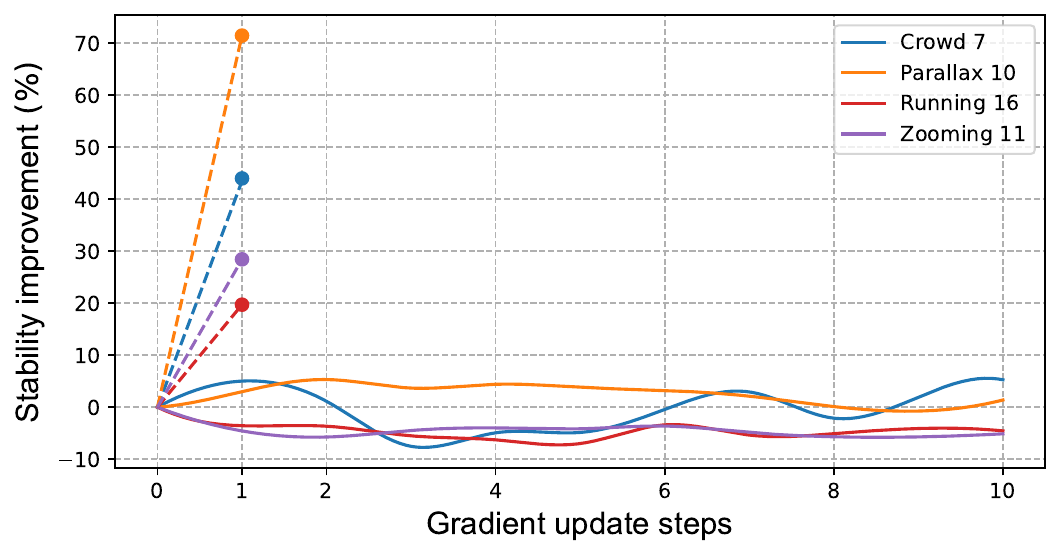}
    \\[-0.5em]
    \caption{
    \textbf{Finetuning vs meta-inference.}
    A comparison of the finetuned and the meta-trained models highlights that it takes significant finetuning iterations for a minuscule improvement.
    Whereas, the proposed algorithm allows for a significant improvement with a single adaptation pass over the video sequence.
    }
    \label{fig:test_time_adaptation_experiment_fig}
    \vspace{-20px}
\end{figure}

\begin{table}[!h]
  \centering
    \adjustbox{max width=0.48\textwidth}{
    \begin{tabular}{c|cccccc}
    \toprule
    Category & \multicolumn{1}{c}{Crowd} & \multicolumn{1}{c}{Parallax} & \multicolumn{1}{c}{Regular} & \multicolumn{1}{c}{Running} & \multicolumn{1}{c}{Quick Rot} & \multicolumn{1}{c}{Zoom} \\
    \midrule
    $\lambda_s$ & 10    & 1     & 1     & 10    & 10    & 1 \\
    $\lambda_p$ & 1     & 1     & 1     & 1     & 1     & 1 \\
    \bottomrule
    \end{tabular}}%
\caption{\textbf{Category-specific weights ($\lambda_s$ and $\lambda_p$).} This table highlights the category-specific weights for the proposed loss functions for the adaptation step. The various motion profiles from the NUS dataset~\cite{liu2013bundled} can be efficiently stabilized by employing these weights during the adaptation process.}
  \label{tab:cat_spec_weights}%
  \vspace{-10pt}
\end{table}%
\noindent
\textbf{Finteuning VS. meta-training.} To highlight the efficacy of the proposed algorithm, we also conducted an ablation study in which we finetuned the baseline DMBVS~\cite{bmvc_ver} with the proposed inner-loop losses on its worst-performing videos (with a stability score of 10$\sim$15\%) from the evaluation dataset and compared the performance of its meta-trained variant with only $1$ adaptation pass (please note that in both of these experiments, we opted the best settings of hyperparameters presented above). We present our findings in Fig.~\ref{fig:test_time_adaptation_experiment_fig}.
The meta-trained model performs significantly well as compared to the baseline. 
\section{Experimental results}\label{res_seciton}
\subsection{Qualitative results}
\begin{figure}[h]
    \centering
    \includegraphics[width=0.98\linewidth]{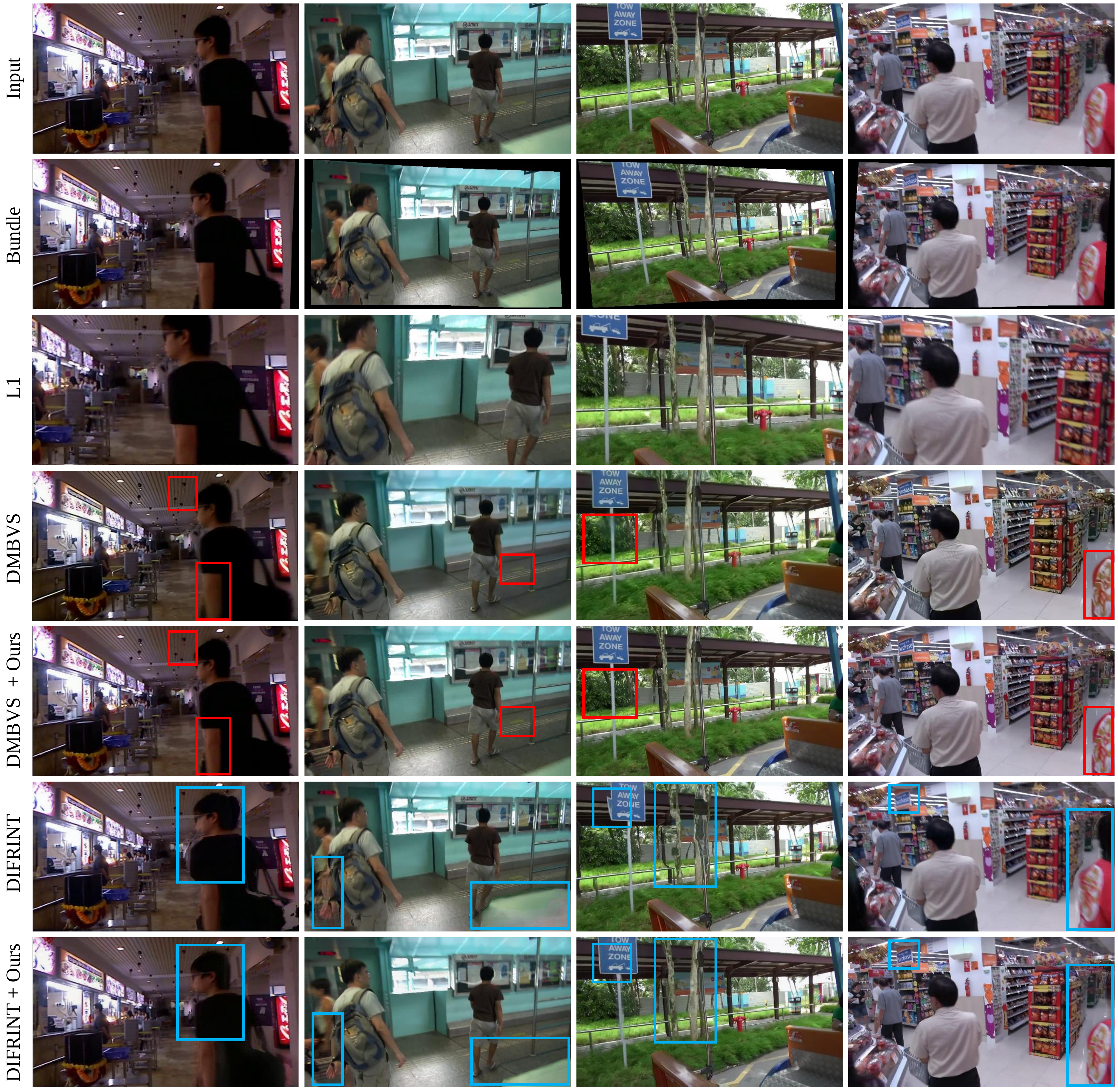}
    \caption{\textbf{Qualitative Results.}
    Qualitative comparison of the meta-trained, baseline models and current SOTA methods.
    The proposed methodology improves the stability of considered models and also mitigates the artifacts present in frame recurrent baseline results.
    \textbf{(Best viewed on a computer screen with zoom).}}
    \label{res:qual_comp}
    \vspace{-18pt}
\end{figure}

For qualitative comparison, we compare our results with L1 stabilizer~\cite{grundmann2011auto}, bundled, and baselines~\cite{bmvc_ver, DIF} in Fig.~\ref{res:qual_comp}.
The bounded regions highlight the temporal artifacts present in DIFRINT~\cite{DIF} and the frame recurrent extension of DMBVS~\cite{bmvc_ver}.
The proposed algorithm mitigates these temporal artifacts successfully and produces sharper results. Due to the space limitation, we only present the qualitative comparison with the longstanding SOTA methods in the main paper and humbly request the readers to refer to the accompanied supplemental for qualitative comparison with other approaches used for quantitative comparison.
\subsection{Quantitative results}
We compare the quantitative performance of both scene-adaptive models with their baseline variants on the NUS dataset~\cite{liu2013bundled} in terms of stability, cropping, and distortion\footnote{Please refer to the accompanied supplemental for the implementation details of these metrics.} in Tab.~\ref{tab:comp_with_baselines}.
This dataset contains videos of $6$ distinct categories including different motion profiles.
The test-time adapted models perform significantly better than their baseline (non-adaptive) counterparts.
We see an average of $5\%$ gain in absolute stability with a single adaptation pass on the test videos for the meta-trained variant of DMBVS~\cite{bmvc_ver}, and an average gain of $8\%$ for DIFRINT~\cite{DIF}.
Please note that this gain does not come at the cost of compromising the full-frame nature of the baseline models and an improvement is also observed in terms of the distortion score as well as evident from Tab.~\ref{tab:comp_with_sotas}.
\begin{table}[t]
\adjustbox{max width=0.48\textwidth}{
  \centering
  
    \begin{tabular}{c|c|cccccc}
    \toprule
    \multicolumn{2}{c|}{\multirow{2}[4]{*}{Model}} & \multicolumn{6}{c}{Stability} \\
\cmidrule{3-8}    \multicolumn{2}{c|}{} & Crowd & Parallax & Regular & Running & Quick Rot & Zoom \\
    \midrule
    \multirow{3}[1]{*}{DMBVS~\cite{bmvc_ver}} & Baseline & \cellcolor[rgb]{1.0, 1.0, 1.0}0.7315 & \cellcolor[rgb]{1.0, 1.0, 1.0}0.7660 & \cellcolor[rgb]{1.0, 1.0, 1.0}0.6938 & \cellcolor[rgb]{1.0, 1.0, 1.0}0.6522 & \cellcolor[rgb]{1.0, 1.0, 1.0}0.8453 & \cellcolor[rgb]{1.0, 1.0, 1.0}0.7811 \\
          & $\text{Adapt}^{\text{(1)}}_{100}$ & \cellcolor[rgb]{ .608,  .761,  .902}0.7584 & \cellcolor[rgb]{ .608,  .761,  .902}0.7965 & \cellcolor[rgb]{ .608,  .761,  .902}0.7133 & \cellcolor[rgb]{ .608,  .761,  .902}0.6983 & \cellcolor[rgb]{ .608,  .761,  .902}0.8906& \cellcolor[rgb]{ .608,  .761,  .902}0.8368 \\
          & $\text{Adapt}^{\text{(5)}}_{100}$ & \cellcolor[rgb]{ .663,  .816,  .557}0.7616 & \cellcolor[rgb]{ .663,  .816,  .557}0.8125 & \cellcolor[rgb]{ .663,  .816,  .557}0.7290 & \cellcolor[rgb]{ .663,  .816,  .557}0.7144 & \cellcolor[rgb]{ .663,  .816,  .557}0.9046 & \cellcolor[rgb]{ .663,  .816,  .557}0.8412 \\
    \midrule
    \multirow{3}[1]{*}{DIFRINT~\cite{DIF}} & Baseline & \cellcolor[rgb]{1.0, 1.0, 1.0}0.7453 & \cellcolor[rgb]{1.0, 1.0, 1.0}0.8321 & \cellcolor[rgb]{1.0, 1.0, 1.0}0.6371 & \cellcolor[rgb]{1.0, 1.0, 1.0}0.7143 & \cellcolor[rgb]{1.0, 1.0, 1.0}0.9058 & \cellcolor[rgb]{1.0, 1.0, 1.0}0.8258 \\
          & $\text{Adapt}^{\text{(1)}}_{100}$ & \cellcolor[rgb]{ .608,  .761,  .902}0.8062 & \cellcolor[rgb]{ .608,  .761,  .902}0.8492 & \cellcolor[rgb]{ .608,  .761,  .902}0.6501 & \cellcolor[rgb]{ .608,  .761,  .902}0.7218 & \cellcolor[rgb]{ .608,  .761,  .902}0.9361 & \cellcolor[rgb]{ .608,  .761,  .902}0.8501 \\
          & $\text{Adapt}^{\text{(5)}}_{100}$ & \cellcolor[rgb]{ .663,  .816,  .557}0.8149 & \cellcolor[rgb]{ .663,  .816,  .557}0.8542 & \cellcolor[rgb]{ .663,  .816,  .557}0.6617 & \cellcolor[rgb]{ .663,  .816,  .557}0.7410 & \cellcolor[rgb]{ .663,  .816,  .557}0.9431 & \cellcolor[rgb]{ .663,  .816,  .557}0.8611 \\
    \bottomrule
    \end{tabular}%
    }
    \caption{\textbf{Quantitative comparison of adapted models against baselines.}
    The proposed algorithm consistently improves the stability with the increasing number of adaptation iterations for both of the considered models.
    The subscript shows the number of sequences sampled for adaptation and the superscript denotes the adaptation number.
    The best stability is highlighted with a green color and the second best is highlighted with a blue color.}
  \label{tab:comp_with_baselines}%
  \vspace{-15pt}
\end{table}%
After our baseline comparison, we present a thorough quantitative assessment against well-established SOTA methods known for their stability~\cite{grundmann2011auto, liu2013bundled}, recent methods~\cite{liu2021hybrid, wang2018deep, yu2020learning, Zhang_2023_ICCV}, and Adobe Premiere Pro 2020's professionally used warp stabilizer in Tab.~\ref{tab:comp_with_sotas}.
Despite the classical nature of the methodologies introduced in~\cite{grundmann2011auto, liu2013bundled}, these approaches still produce state-of-the-art results, in terms of stability~\cite{VS_survey}. Please note that the proposed method in~\cite{Zhang_2023_ICCV} produces video results across the entire evaluation dataset, however, it is imperative to highlight that videos generated by this method exhibit pronounced shakes in the initial frames, gradually leading to stable videos due to their inherent minimum latency constraints. This instability in the initial segment (spanning over 30 frames per video) impedes the estimation of homography for stability metric calculation. To ensure a fair comparison, we only present average results from their method where the stability metric can be computed for the entire videos.

\begin{table}[t]
  \centering
    \resizebox{\columnwidth}{!}{\begin{tabular}{l|c|c|c}
    \toprule
    Method & Stability $\uparrow$ & \multicolumn{1}{l|}{Cropping $\uparrow$} & \multicolumn{1}{l}{Distortion $\uparrow$} \\
    \midrule
    L1~\cite{grundmann2011auto}    & 0.8661 & 0.7392 & 0.9215 \\
    Bundled~\cite{liu2013bundled} & \cellcolor[rgb]{ .608,  .761,  .902}0.8750 & 0.8215 & 0.7781 \\
    Adobe Premiere Pro 2020*    & 0.8262 & 0.7432 & 0.8230 \\
    StabNet*~\cite{wang2018deep}    & 0.7422 & 0.6615 & 0.8878 \\
    Yu and Ramamoorthi~\etal~\cite{yu2020learning}    & 0.7905 & 0.8592 & 0.9105 \\
    FuSta~\cite{liu2021hybrid}    & 0.8037 & 0.9992 & 0.9642 \\
    Zhang et al.*~\cite{Zhang_2023_ICCV} & 0.7481 & 0.9592 & \cellcolor[rgb]{ .608,  .761,  .902}0.9988 \\
    DMBVS~\cite{bmvc_ver} (baseline) & 0.7372 & 0.9983 & 0.9189 \\
    DMBVS~\cite{bmvc_ver} + $\text{Adapt}^{\text{(1)}}_{100}$ & 0.7532 & 0.9974 & 0.9112 \\
    DMBVS~\cite{bmvc_ver} + $\text{Adapt}^{\text{(5)}}_{100}$ & 0.7852 & 0.9973 & 0.9461 \\
    DMBVS~\cite{bmvc_ver} + $\text{Adapt}^{\text{(1)}}_{\text{all}}$ & 0.7760 & \cellcolor[rgb]{ .663,  .816,  .557}0.9999 & \cellcolor[rgb]{ .663,  .816,  .557}0.9990 \\
    DMBVS~\cite{bmvc_ver} + $\text{Adapt}^{\text{(1)}}_{all}$ + recurrent & 0.7867 & 0.9999 & \cellcolor[rgb]{ .608,  .761,  .902}0.9818 \\
    DIFRINT~\cite{DIF} (baseline) & 0.7904 & 0.9993 & 0.9438 \\
    DIFRINT~\cite{DIF} + $\text{Adapt}^{\text{(1)}}_{100}$ & 0.8428 & 0.9993 & 0.9587 \\
    DIFRINT~\cite{DIF} + $\text{Adapt}^{\text{(5)}}_{100}$ & 0.8528 & \cellcolor[rgb]{ .608,  .761,  .902}0.9994 & 0.9596 \\
    DIFRINT~\cite{DIF} + $\text{Adapt}^{\text{(1)}}_{all}$ & \cellcolor[rgb]{ .663,  .816,  .557}0.8786 & \cellcolor[rgb]{ .608,  .761,  .902}0.9994 & 0.9569 \\
    \bottomrule
    
    \end{tabular}%
    }
    \caption{\textbf{Quantitative Results.}
    The proposed algorithm consistently improves the stability with the increasing number of adaptation iterations for both of the considered models.
    The proposed algorithm enables DIFRINT~\cite{DIF} to achieve SOTA results with a single adaptation iteration over all the frame sequences in videos from the NUS dataset~\cite{liu2013bundled}.
    Please note that the methods proposed in~\cite{wang2018deep} and Adobe Premiere Pro fail to stabilize some videos; therefore, their results are averaged over only the stabilized videos.}
  \label{tab:comp_with_sotas}
\vspace{-16pt}
\end{table}

The proposed algorithm consistently improves the results of both the considered models and equips DIFRINT to achieve SOTA results and also improves the mean stability of DMBVS without compromising the full-frame nature or quality of the stabilized videos. 

Please note that the average stability of the adapted method can be further increased by opting for a higher number of adaptation iterations and higher weights for the stability losses during the adaptation process.
In Tab.~\ref{tab:comp_with_baselines}, we only present the results generated with up to a single adaptation iteration on each consecutive sequence due to the time complexity and resource limitations.
In order to significantly cut down the time required for adaptation, we observe that comparable results can be achieved by adapting on a constant number of randomly sampled sequences with a higher number of adaptation iterations (as evident from Tab.~\ref{tab:comp_with_sotas}).
Furthermore, the quality of the results (as indicated by the Distortion metric) also suggests that the proposed algorithm not only improves the stability but consistently enhances the quality as well.
Please note that employing the iterative strategy proposed in~\cite{DIF} can further enhance the stability of the resultant videos.
Please refer to the accompanied supplemental for user studies and other metric results.

\section{Conclusion}
In this study, we aim to improve full-frame pixel-level synthesis video stabilization solutions by leveraging additional information available at test time. We introduce a meta-learning algorithm for this task, enabling rapid adaptation of model parameters for scenes containing unique motion profiles. Our proposed algorithm's versatility is demonstrated through extensive experimentation on publicly available models for this task. The proposed algorithm enables the users to control various aspects of video stabilization (to an extent), which was previously unattainable for such models, and shows consistent improvement in both stability and quality. The proposed algorithm can be seamlessly integrated with upcoming pixel-synthesis solutions for this task without additional parametric or structural changes.
\clearpage
\section*{Acknowledgements}
This work was supported by Institute of Information \& communications Technology Planning \& Evaluation (IITP) grant funded by the Korea government(MSIT) (No.2022- 0-00156, Fundamental research on continual meta-learning for quality enhancement of casual videos and their 3D metaverse transformation) and Samsung Electronics Co., Ltd, and Samsung Research Funding Center of Samsung Electronics under Project Number SRFCIT1901-06.
{
    \small
    \bibliographystyle{ieeenat_fullname}
    \bibliography{main}
}


\end{document}